\documentclass[runningheads]{llncs}
\usepackage[T1]{fontenc}
\usepackage{graphicx}
\usepackage{amsmath}
\usepackage{amssymb}
\usepackage{booktabs}
\usepackage{array}
\begin{document}
\title{Multilingual Sentiment-Aware Text Summarization: A Reinforcement Learning Approach for Consistency Maintenance}
\titlerunning{Multilingual Sentiment-Aware Summarization}
%
\author{Mikhail Krasitskii\inst{1}\orcidID{0009-0005-2581-8110} \and
Alexander Gelbukh\inst{1}\orcidID{0000-0001-7845-9039} \and
Olga Kolesnikova\inst{1}\orcidID{0000-0002-1307-1647} \and
Grigori Sidorov\inst{1}\orcidID{0000-0003-3901-3522}}
\authorrunning{M. Krasitskii et al.}
%
\institute{Instituto Politécnico Nacional (IPN), Centro de Investigación en Computación (CIC), Mexico City, Mexico \\
\email{\{mkrasitskii2023, gelbukh, kolesnikova, sidorov\}@cic.ipn.mx}
%
}
\maketitle              
\begin{abstract}
Reinforcement Learning from Human Feedback (RLHF) has significantly improved the quality and fluency of large language models in text summarization. However, its impact on affective properties remains insufficiently understood. In this work, we study \textit{sentiment drift}, a systematic shift toward neutral sentiment in RLHF-based summarization outputs compared to source texts.

We conduct extensive experiments across multiple datasets, model architectures, and eight languages to analyze how alignment objectives influence sentiment preservation. Our results show that sentiment drift is a consistent phenomenon that becomes stronger with increased KL regularization strength, indicating a trade-off between alignment stability and affective fidelity.

To explain this behavior, we introduce a Policy Attribution framework that decomposes the RLHF objective and quantifies the contribution of its components. Our analysis reveals that KL regularization is the primary driver of sentiment suppression across all settings.

Based on these findings, we propose a sentiment-aware modification of the KL regularization term, which selectively reduces constraints on sentiment-bearing tokens. Empirical results demonstrate that this approach mitigates sentiment drift while maintaining summarization quality.

Overall, our findings highlight a fundamental limitation of current alignment methods: while they improve factual consistency and safety, they may unintentionally suppress emotional expressiveness. This motivates the development of alignment strategies that explicitly account for affective preservation.

\keywords{Text Summarization \and Reinforcement Learning \and Sentiment Analysis \and Multilingual NLP \and RLHF \and Policy Attribution}
\end{abstract}


\section{Introduction}
\label{sec:introduction}

Reinforcement Learning from Human Feedback (RLHF) has become a central paradigm for aligning large language models with human preferences, significantly improving fluency, coherence, and overall output quality~\cite{stiennon2020learning,ouyang2022training}. In text summarization, RLHF enables models to produce outputs that better reflect human judgments of relevance and readability.

However, this improvement often introduces an important trade-off. Aligned models tend to produce more conservative and less expressive outputs, favoring safe and predictable language. This behavior has been linked to reward over-optimization and alignment constraints, which encourage models to remain close to a reference policy~\cite{gao2022scaling}. As a result, generated summaries frequently exhibit reduced variability and diminished affective content.

In this work, we focus on a specific manifestation of this phenomenon, which we refer to as \textit{sentiment drift}. Sentiment drift describes the systematic shift toward neutral tone in generated summaries compared to the source text. While factual content is largely preserved, the emotional and subjective aspects of the original text are often weakened or lost. This effect is particularly relevant for downstream tasks such as sentiment analysis and opinion mining, where preserving affective information is essential.

Despite its practical importance, sentiment drift remains insufficiently understood. Existing research in summarization has primarily focused on content selection and faithfulness~\cite{gehrmann2018bottomup,laban2022summac,maynez2020faithfulness}, while work on sentiment-aware generation typically introduces explicit constraints without analyzing how alignment objectives influence sentiment implicitly~\cite{amplayo2021unsupervised}. Moreover, in multilingual settings, sentiment expression varies across languages, and text compression can introduce additional distortions in sentiment distributions~\cite{balahur2014sentiment,krasitskii2025multilingual}.

We hypothesize that sentiment drift arises directly from the RL optimization objective, particularly from the Kullback–Leibler (KL) regularization term. This term penalizes deviations from a reference policy and disproportionately suppresses low-frequency or high-variance tokens, which often carry sentiment. As a result, sentiment-bearing tokens are more likely to be attenuated during training.

To investigate this hypothesis, we introduce a \textit{Policy Attribution} framework that decomposes the RL objective into its constituent components and enables fine-grained analysis of their contributions to sentiment suppression. This allows us to quantify the role of KL regularization and better understand the mechanisms underlying sentiment drift.

The main contributions of this work are as follows:
\begin{itemize}
    \item We formally define sentiment drift and propose quantitative metrics for its evaluation.
    \item We introduce a Policy Attribution framework for analyzing the contribution of RL objective components.
    \item We provide empirical evidence that KL regularization is a primary driver of sentiment suppression.
    \item We propose a sentiment-aware KL regularization method that mitigates drift while preserving summarization quality.
\end{itemize}

Overall, our findings highlight the need to explicitly account for affective properties in alignment objectives. While current methods improve factual quality and safety, they may unintentionally reduce expressive fidelity, particularly in multilingual settings.


\section{Background and Related Work}
\label{sec:related_work}

\subsection{RLHF and Alignment Methods}

RLHF is a standard approach for aligning large language models with human preferences~\cite{stiennon2020learning,ouyang2022training}. The typical pipeline combines supervised fine-tuning, reward modeling, and reinforcement learning, most commonly implemented using Proximal Policy Optimization (PPO)~\cite{schulman2017ppo}.

The optimization objective in RLHF balances reward maximization with a KL divergence constraint that penalizes deviations from a reference policy. While this constraint improves training stability and output quality, it also encourages conservative generation behavior, as the model is discouraged from producing outputs that differ significantly from the reference distribution.

Alternative alignment approaches, such as Direct Preference Optimization (DPO)~\cite{rafailov2023direct}, directly optimize preference data without explicit reward modeling. Despite differences in formulation, both RLHF and DPO impose alignment constraints that may influence not only content quality but also stylistic and affective properties of generated text.

\subsection{Conservative Generation and Reward Optimization}

Prior work has shown that reward-based optimization can lead to over-optimization effects, where models exploit reward signals and converge to safer, less diverse outputs~\cite{gao2022scaling}. This phenomenon is often associated with conservative generations, where models prefer predictable and high-probability sequences.

Although this behavior has been studied primarily in the context of reward hacking and factual consistency, its impact on sentiment and expressive properties remains underexplored. In particular, the interaction between alignment constraints and affective content has not been systematically analyzed.

\subsection{Sentiment Preservation in Summarization}

Abstractive summarization research has traditionally focused on content selection and fluency~\cite{gehrmann2018bottomup}, as well as factual consistency and faithfulness~\cite{laban2022summac,maynez2020faithfulness}. More recent work highlights the importance of preserving subjective and opinionated content, especially in tasks involving user-generated text~\cite{amplayo2021unsupervised}.

However, most sentiment-aware approaches introduce explicit modeling constraints and do not analyze how alignment objectives implicitly affect sentiment during training. This leaves an important gap in understanding how modern alignment methods influence affective properties of generated summaries.

In multilingual settings, sentiment expression varies across languages, and translation or compression can introduce systematic shifts in sentiment distributions~\cite{balahur2014sentiment,krasitskii2025multilingual}. These effects suggest that sentiment preservation is inherently more complex in cross-lingual scenarios.

\subsection{Attribution for Model Analysis}

Attribution methods such as Integrated Gradients provide a principled way to analyze neural network behavior by decomposing model outputs into contributions of individual inputs~\cite{sundararajan2017axiomatic}. These techniques have been widely used to study model decisions and feature importance.

However, existing applications primarily focus on correctness and relevance, with limited attention to stylistic or affective properties. In particular, there is a lack of methods that quantify how different components of training objectives influence sentiment-related behavior in large language models.

\paragraph{Summary.} Overall, prior work has not systematically analyzed how alignment objectives affect sentiment preservation, nor quantified the contribution of individual components such as KL regularization. This gap is particularly pronounced in multilingual settings, motivating the need for a more detailed and mechanistic analysis.


\section{Methodology}
\label{sec:methodology}

This section presents the methodological framework for analyzing sentiment drift in RLHF-based summarization. We formalize the optimization objective, define sentiment evaluation metrics, introduce a Policy Attribution framework, and describe the proposed sentiment-aware regularization.

\subsection{RLHF Objective}

We consider a standard RLHF setup for abstractive summarization~\cite{stiennon2020learning,ouyang2022training}. Given an input text $x$, a policy $\pi_{\theta}$ generates a summary $y = (y_1, \dots, y_T)$ and is optimized to maximize a reward model $R_{\phi}(x, y)$ while remaining close to a reference policy $\pi_{\text{ref}}$.

The optimization objective is defined as:

\begin{equation}
    J(\theta) = \mathbb{E}_{x,y \sim \pi_{\theta}} \left[
    R_{\phi}(x, y) - \beta \, D_{\text{KL}} \bigl(\pi_{\theta}(\cdot|x) \parallel \pi_{\text{ref}}(\cdot|x)\bigr)
    \right],
\end{equation}

where $\beta > 0$ controls the strength of KL regularization.\\

In practice, optimization is performed using PPO~\cite{schulman2017ppo}, which minimizes the surrogate loss:

\begin{equation}
    \mathcal{L}(\theta) = \mathcal{L}_{\text{reward}} + \beta \mathcal{L}_{\text{KL}} + \mathcal{L}_{\text{clip}},
\end{equation}

Here, $\mathcal{L}_{\text{reward}}$ corresponds to the negative expected reward, $\mathcal{L}_{\text{KL}}$ enforces proximity to the reference policy, and $\mathcal{L}_{\text{clip}}$ ensures stable policy updates. This establishes a direct correspondence between the maximization of $J(\theta)$ and the minimization of $\mathcal{L}(\theta)$.

\subsection{Sentiment Evaluation Metrics}

To quantify sentiment drift, we compare sentiment distributions between source texts $S$ and generated summaries $\hat{S}$.

\paragraph{Sentiment Variance (SV).}
SV measures the dispersion of sentiment scores within a text. Lower SV indicates reduced emotional expressiveness and a shift toward neutrality. Importantly, lower SV does not necessarily imply better performance, but rather stronger neutrality bias.

\paragraph{Jensen-Shannon Divergence (JSD).}
We measure the divergence between sentiment distributions of $S$ and $\hat{S}$ using:

\begin{equation}
    \text{JSD}(P \parallel Q) = \frac{1}{2} D_{\text{KL}}(P \parallel M) + \frac{1}{2} D_{\text{KL}}(Q \parallel M),
    \label{eq:jsd}
\end{equation}

where $M = \frac{1}{2}(P + Q)$. Lower JSD indicates better sentiment preservation.

\subsection{Policy Attribution Framework}

To analyze how different components of the RL objective influence sentiment, we decompose the training loss gradient as:

\begin{equation}
    \nabla_{\theta} \mathcal{L} = 
    \nabla_{\theta} \mathcal{L}_{\text{reward}} +
    \beta \nabla_{\theta} \mathcal{L}_{\text{KL}} +
    \nabla_{\theta} \mathcal{L}_{\text{clip}}.
\end{equation}

We estimate component-wise contributions using Integrated Gradients~\cite{sundararajan2017axiomatic}. For a given token $t_i$, the attribution score for component $c \in \{\text{reward}, \text{KL}, \text{clip}\}$ is defined as:

\begin{equation}
    A_c(t_i) = (\theta - \theta') \cdot 
    \int_{0}^{1}
    \frac{\partial \mathcal{L}_c(\theta' + \alpha(\theta - \theta'))}
    {\partial \log \pi_{\theta}(t_i | x, y_{<i})}
    \, d\alpha,
\end{equation}

where $\theta'$ denotes the baseline parameters corresponding to the supervised fine-tuned (SFT) model prior to RL optimization.

In practice, the integral is approximated using a Riemann sum with $K$ interpolation steps:

\begin{equation}
    A_c(t_i) \approx \frac{1}{K} \sum_{k=1}^{K}
    (\theta - \theta') \cdot
    \frac{\partial \mathcal{L}_c(\theta_k)}
    {\partial \log \pi_{\theta_k}(t_i | x, y_{<i})},
\end{equation}

where $\theta_k = \theta' + \frac{k}{K}(\theta - \theta')$.

Negative attribution values for the KL component indicate suppression of token probabilities during training.

\subsection{Identification of Sentiment-Bearing Tokens}

We identify sentiment-bearing tokens using external multilingual sentiment classifiers. Specifically, sentence-level sentiment predictions are first obtained, after which token-level relevance is estimated using gradient-based attribution with respect to the classifier output.

A binary indicator function is defined as:

\begin{equation}
    \mathbf{1}_{\text{sent}}(t_i) =
        \begin{cases}
            1, & \text{if } t_i \text{ contributes to sentiment polarity}, \\
            0, & \text{otherwise}.
        \end{cases}
\end{equation}

This approach allows consistent identification of sentiment-bearing tokens across languages while partially accounting for contextual dependencies beyond isolated tokens.

\subsection{Sentiment-Aware KL Regularization}

To mitigate sentiment drift, we modify the KL regularization term by introducing token-level weights:

\begin{equation}
    J_{\text{SA}}(\theta) =
    \mathbb{E}_{x,y} \left[
    R_{\phi}(x, y) - \beta \sum_{i=1}^{|y|} w_i \, \text{KL}_i
    \right],
    \label{eq:sa_kl}
\end{equation}

where weights are defined as:

\begin{equation}
    w_i = 1 - \gamma \cdot \mathbf{1}_{\text{sent}}(t_i),
\end{equation}

and $\gamma \in [0,1]$ controls the degree of relaxation.

This formulation selectively reduces the KL penalty for tokens contributing to sentiment, while preserving the overall regularization structure and stability of training.

\subsection{Computational Considerations}

To ensure computational feasibility for large models (7B–8B parameters), attribution is computed on a subset of tokens corresponding to the top 10\% highest sentiment relevance scores. The number of interpolation steps is fixed to $K=50$.

Attribution scores are aggregated across tokens, samples, and batches to estimate component-level contributions, which are later reported in the Results section.


\section{Experimental Setup}
\label{sec:experimental_setup}

We conduct controlled experiments to evaluate sentiment drift in RLHF-based summarization using standard models and datasets.

\subsection{Datasets}

We use two widely adopted summarization benchmarks:

\begin{itemize}
    \item \textbf{Reddit TL;DR}~\cite{stiennon2020learning}: informal texts with strong sentiment signals, suitable for analyzing sentiment preservation.
    \item \textbf{CNN/DailyMail}~\cite{gehrmann2018bottomup}: news articles with more neutral and structured language, providing a complementary evaluation setting.
\end{itemize}

We construct train/validation/test splits as follows:
\begin{itemize}
    \item Reddit TL;DR: 100K / 5K / 5K samples
    \item CNN/DailyMail: 287K / 13K / 11K samples
\end{itemize}

To ensure reliable sentiment evaluation, we retain only samples with clear sentiment polarity using a confidence threshold of 0.6 from a pre-trained multilingual sentiment classifier (XLM-R based).

\paragraph{Multilingual Setup.} Datasets are translated into eight languages (EN, ES, DE, FR, IT, AR, FI, HU) using the NLLB-200 (3.3B) model. Sentence-level alignment is preserved, and samples with inconsistent sentiment polarity after translation are filtered out. Translation quality is verified using round-trip consistency checks on a random subset (1K samples per language).

\subsection{Model Configuration}

Experiments are conducted using LLaMA-3-8B and Mistral-7B models. We follow a standard RLHF pipeline implemented with TRL, including supervised fine-tuning, reward modeling, and PPO optimization~\cite{stiennon2020learning,ouyang2022training}.

The KL coefficient is varied across $\{0.05, 0.1, 0.2\}$, while other parameters are fixed to ensure controlled comparison.

\subsection{Evaluation Metrics}

We evaluate both summarization quality and sentiment preservation using:

\begin{itemize}
    \item \textbf{Sentiment Variance (SV)}: measures dispersion of sentiment scores. Lower values indicate stronger neutrality and reduced emotional expressiveness (not necessarily better performance).
    \item \textbf{Jensen-Shannon Divergence (JSD)}: measures distributional differences between source and summary sentiment distributions (Eq.~\ref{eq:jsd}). Lower values indicate better sentiment preservation.
    \item \textbf{ROUGE-L}: evaluates content overlap with reference summaries, capturing factual consistency.
\end{itemize}

\begin{figure}[h!]
    \centering
    \includegraphics[width=0.65\linewidth]{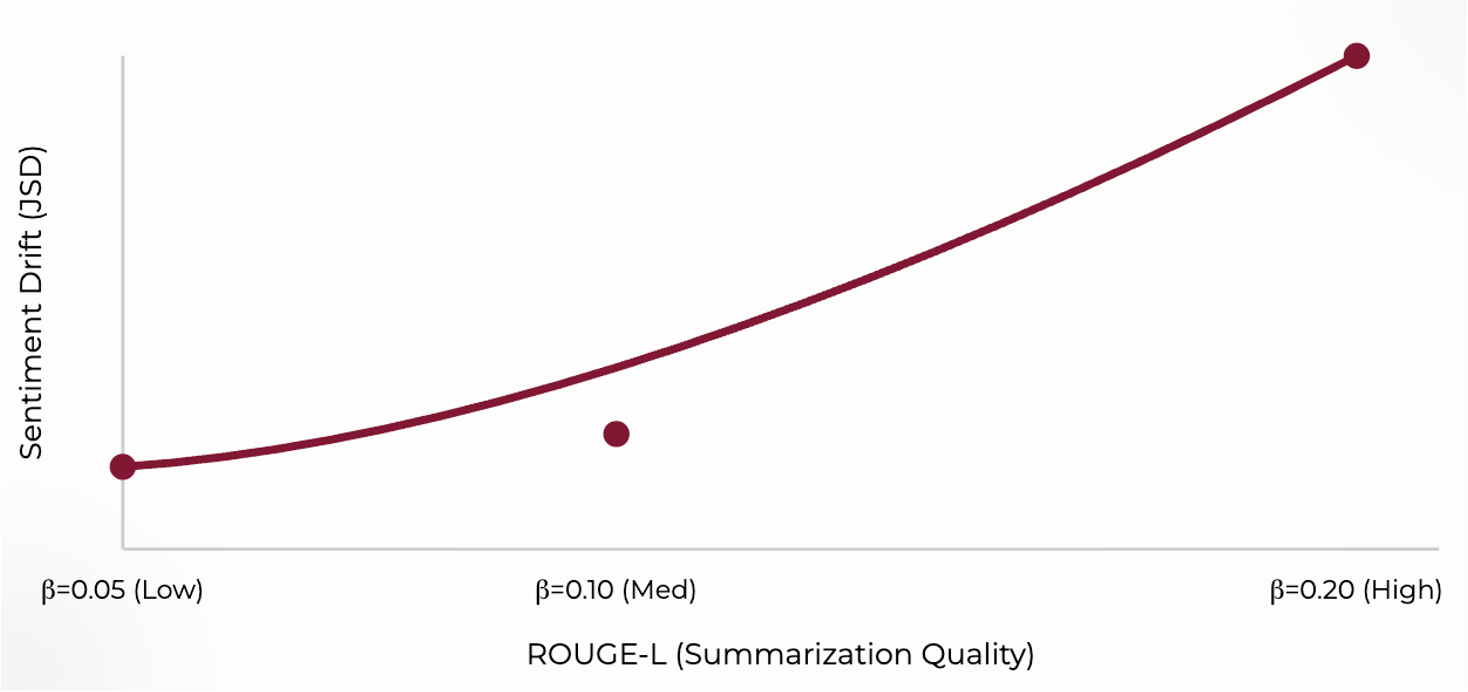}
    \caption{Trade-off between summarization quality (ROUGE-L) and sentiment preservation (JSD). Each point corresponds to a different KL coefficient $\beta$.}
    \label{fig:pareto_tradeoff}
\end{figure}

Figure~\ref{fig:pareto_tradeoff} illustrates the trade-off between summarization quality and sentiment preservation.

\begin{figure}[h!]
    \centering
        \includegraphics[width=0.99\linewidth]{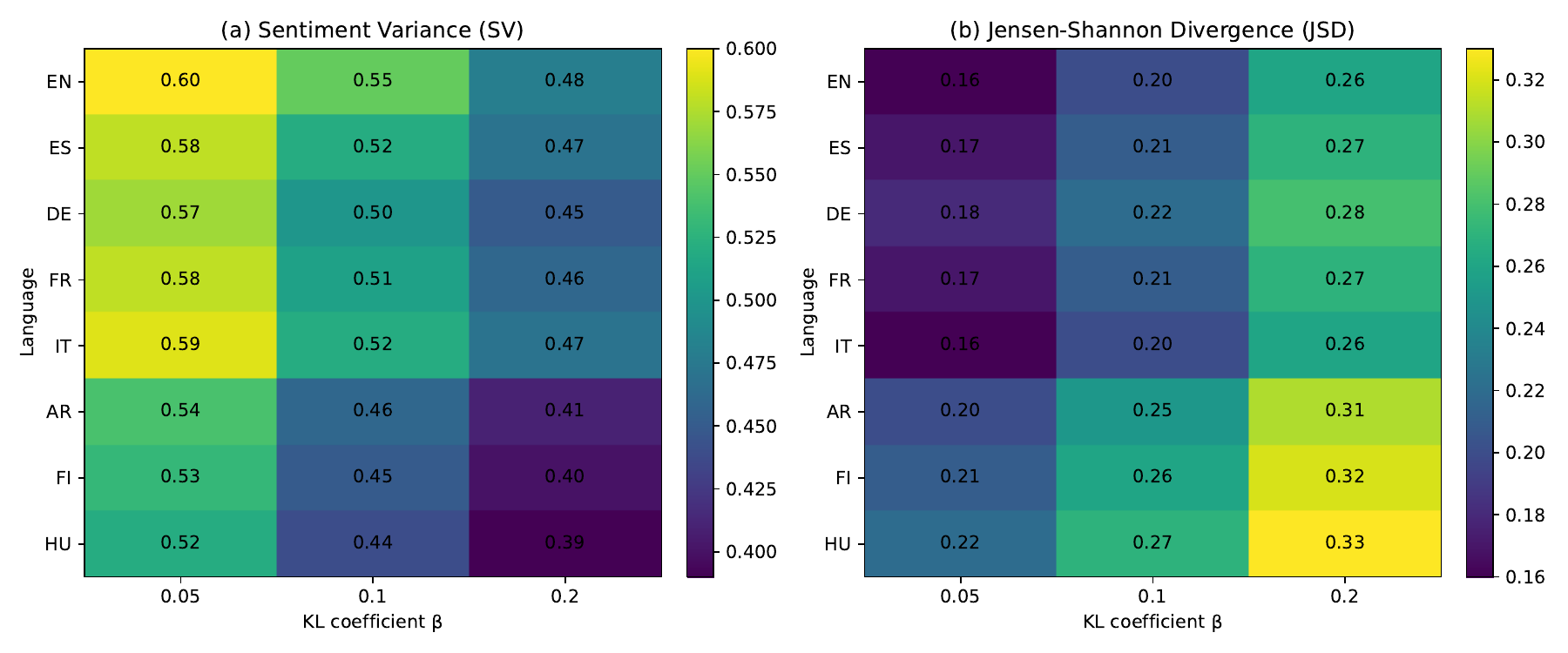}
            \caption{Cross-lingual sentiment drift across KL coefficients $\beta$.
            (a) SV decreases with $\beta$, indicating stronger neutrality bias.
            (b) JSD increases with $\beta$, indicating reduced sentiment preservation.}
    \label{fig:cross_lingual_heatmap}
\end{figure}

Figure~\ref{fig:cross_lingual_heatmap} shows cross-lingual sentiment drift patterns.

\subsection{Policy Attribution Implementation}

We compute attribution scores using Integrated Gradients~\cite{sundararajan2017axiomatic} with 50 interpolation steps. Attribution is computed for the top 10\% of tokens ranked by sentiment relevance.

Scores are aggregated across tokens and samples to estimate the contribution of each objective component. The resulting attribution statistics are reported in Tables~\ref{tab:gradient_attr} and~\ref{tab:attribution} in Section~\ref{sec:results}.

\subsection{RLHF Training Setup}

The RLHF pipeline follows standard practice~\cite{ouyang2022training} and includes:

\begin{itemize}
    \item Supervised Fine-Tuning (SFT) on reference summaries
    \item Reward model trained on preference pairs derived from human annotations
    \item PPO-based optimization
\end{itemize}

\textbf{Hyperparameters:}
\begin{itemize}
    \item learning rate: $1 \times 10^{-5}$
    \item batch size: 64
    \item PPO epochs: 4
    \item clip range: 0.2
    \item optimizer: AdamW
\end{itemize}

\textbf{Decoding:}
\begin{itemize}
    \item temperature = 0.7
    \item top-$p$ = 0.9
    \item max length = 128
\end{itemize}

\paragraph{Hardware.}
All experiments are conducted on NVIDIA A100 GPUs (80GB). Training time per model is approximately 18 hours.

\subsection{DPO Setup}

To assess generality, we conduct additional experiments with DPO~\cite{rafailov2023direct}.

Models are trained on the same preference pairs with $\beta \in \{0.1, 0.2, 0.5\}$ using:

\begin{equation}
L_{\text{DPO}}(\theta) = - \mathbb{E}_{(x, y_w, y_l)} \left[
\log \sigma \left(
\beta \log \frac{\pi_\theta(y_w|x)}{\pi_{\text{ref}}(y_w|x)} -
\beta \log \frac{\pi_\theta(y_l|x)}{\pi_{\text{ref}}(y_l|x)}
\right)
\right]
\label{eq:dpo}
\end{equation}

\subsection{Sentiment-Aware Regularization}

We evaluate the proposed sentiment-aware KL regularization using Eq.~\ref{eq:sa_kl} with $\gamma \in \{0.3, 0.5\}$. All other parameters remain unchanged.

\subsection{Statistical Protocol}

All experiments are run with 3 random seeds. We report mean values and standard deviations. Statistical significance is evaluated using paired t-tests ($p < 0.05$).


\section{Results and Analysis}
\label{sec:results}

Lower SV values indicate stronger neutrality (reduced sentiment expressiveness), while lower JSD values indicate better sentiment preservation.

\subsection{Multilingual Sentiment Drift}

Table~\ref{tab:multilingual_results} presents sentiment preservation results across eight languages. RLHF consistently reduces sentiment variance compared to source texts, indicating a systematic shift toward more neutral outputs.

At the same time, JSD values remain substantial across all languages, showing that generated summaries deviate from original sentiment distributions. Together, these results confirm the presence of sentiment drift.

The magnitude of drift varies across languages. Morphologically rich languages such as Arabic, Finnish, and Hungarian exhibit larger reductions in SV and higher JSD values. This suggests that increased morphological complexity amplifies the suppression of sentiment-bearing tokens under alignment constraints.

This trend is consistent with the cross-lingual patterns shown in Figure~\ref{fig:cross_lingual_heatmap}.

\begin{table}[!htb]
\centering
\caption{Multilingual sentiment preservation results.}
\label{tab:multilingual_results}
\begin{tabular}{|l|c|c|c|c|}
\hline
\textbf{Language} & \textbf{SV Source} & \textbf{SV RLHF $\downarrow$} & \textbf{JSD $\downarrow$} & \textbf{ROUGE-L $\uparrow$} \\
\hline
English & 0.72 & 0.54 & 0.18 & 42.1 \\
\hline
Spanish & 0.70 & 0.52 & 0.20 & 41.3 \\
\hline
German & 0.68 & 0.49 & 0.22 & 40.5 \\
\hline
French & 0.69 & 0.50 & 0.21 & 40.8 \\
\hline
Italian & 0.71 & 0.51 & 0.19 & 41.0 \\
\hline
Arabic & 0.65 & 0.43 & 0.27 & 38.9 \\
\hline
Finnish & 0.66 & 0.42 & 0.29 & 38.5 \\
\hline
Hungarian & 0.64 & 0.40 & 0.31 & 37.9 \\
\hline
\end{tabular}
\end{table}

\subsection{Effect of KL Regularization}

Table~\ref{tab:kl_ablation} shows the effect of varying the KL coefficient $\beta$. Increasing $\beta$ leads to a monotonic decrease in SV and a corresponding increase in JSD, indicating stronger neutrality bias and reduced sentiment preservation.

At the same time, ROUGE-L slightly improves with higher $\beta$, suggesting that stronger regularization promotes conservative generation that favors factual consistency over expressive variation.

This trade-off is illustrated in Figure~\ref{fig:pareto_tradeoff}.

\begin{table}[!htb]
\centering
\caption{Effect of KL coefficient $\beta$ on sentiment drift.}
\label{tab:kl_ablation}
\begin{tabular}{|l|c|c|c|}
\hline
\textbf{$\beta$} & \textbf{SV $\downarrow$} & \textbf{JSD $\uparrow$} & \textbf{ROUGE-L $\uparrow$} \\
\hline
0.05 & 0.58 & 0.17 & 41.9 \\
\hline
0.10 & 0.51 & 0.22 & 42.3 \\
\hline
0.20 & 0.44 & 0.28 & 42.6 \\
\hline
\end{tabular}
\end{table}

\subsection{Attribution Analysis}

To identify the mechanisms behind sentiment suppression, we analyze gradient attribution across objective components.

\begin{figure}[!htb]
\centering
\includegraphics[width=0.85\linewidth]{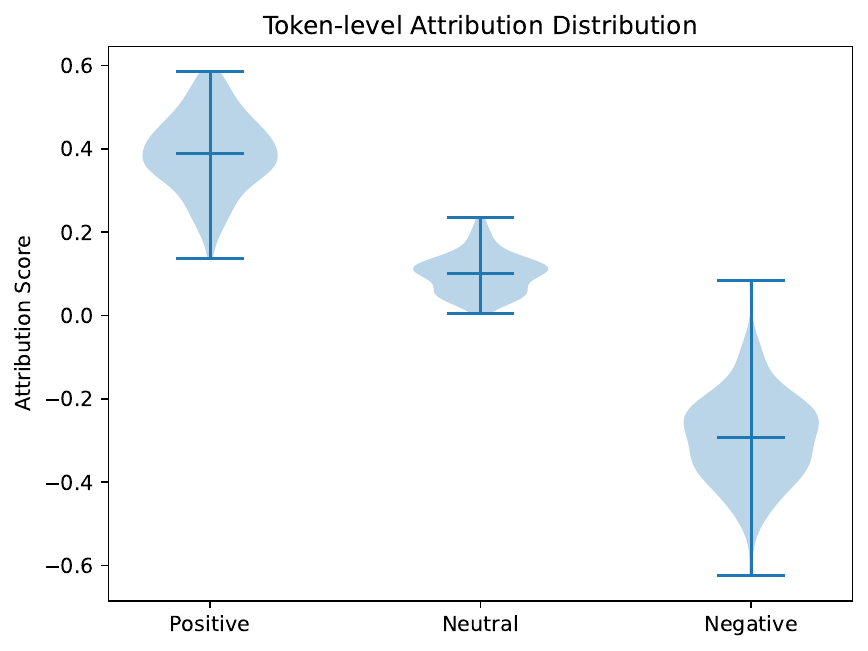}
\caption{Distribution of token-level attribution scores across objective components. Negative values indicate suppression effects.}
\label{fig:attribution_distribution}
\end{figure}

Figure~\ref{fig:attribution_distribution} shows the distribution of token-level attribution scores. The KL component consistently produces negative contributions, indicating systematic suppression of token probabilities.

Table~\ref{tab:gradient_attr} confirms this trend across sentiment categories.

\begin{table}[!htb]
\centering
\caption{Gradient attribution scores by objective component.}
\label{tab:gradient_attr}
\begin{tabular}{|l|c|c|c|}
\hline
\textbf{Component} & \textbf{Positive} & \textbf{Neutral} & \textbf{Negative} \\
\hline
Reward & 0.42 & 0.35 & 0.40 \\
\hline
KL & -0.31 & -0.12 & -0.34 \\
\hline
Clip & 0.05 & 0.02 & 0.04 \\
\hline
\end{tabular}
\end{table}

Aggregated results in Table~\ref{tab:attribution} show that KL accounts for 62\% of total suppression, confirming it as the dominant driver of sentiment drift.

\begin{table}[!htb]
\centering
\caption{Aggregated attribution over sentiment-bearing tokens.}
\label{tab:attribution}
\begin{tabular}{|l|c|}
\hline
\textbf{Component} & \textbf{Suppression Ratio} \\
\hline
KL & 0.62 \\
\hline
Reward & 0.28 \\
\hline
Clip & 0.10 \\
\hline
\end{tabular}
\end{table}

\subsection{Comparison with DPO}

Table~\ref{tab:dpo_results} compares PPO-based RLHF with DPO. Both methods reduce SV relative to the SFT baseline, indicating a general neutrality bias in preference-based alignment.

However, DPO exhibits slightly weaker sentiment drift (higher SV, lower JSD) than PPO, suggesting that the effect is not tied to a specific optimization algorithm but is inherent to preference optimization frameworks.

\begin{table}[!htb]
\centering
\caption{Comparison between PPO-based RLHF and DPO.}
\label{tab:dpo_results}
\begin{tabular}{|l|c|c|c|c|}
\hline
\textbf{Method} & \textbf{SV $\downarrow$} & \textbf{JSD $\downarrow$} & \textbf{ROUGE-L $\uparrow$} & \textbf{Drift Strength} \\
\hline
SFT Baseline & 0.210 & 0.082 & 35.8 & Low \\
\hline
RLHF (PPO) & 0.121 & 0.145 & 36.8 & High \\
\hline
DPO & 0.138 & 0.132 & 36.6 & Medium \\
\hline
\end{tabular}
\end{table}

\subsection{Mitigation via Sentiment-Aware Regularization}

Table~\ref{tab:sakl_results} shows that the proposed sentiment-aware KL regularization improves sentiment preservation.

Increasing $\gamma$ leads to higher SV and lower JSD compared to standard RLHF, while ROUGE-L remains comparable. This indicates that sentiment drift can be mitigated without degrading summarization quality.

\begin{table}[!htb]
\centering
\caption{Effect of Sentiment-Aware KL Regularization on drift mitigation.}
\label{tab:sakl_results}
\begin{tabular}{|l|c|c|c|c|}
\hline
\textbf{Method} & \textbf{SV $\downarrow$} & \textbf{JSD $\downarrow$} & \textbf{ROUGE-L $\uparrow$} & \textbf{Drift Reduction} \\
\hline
SFT Baseline & 0.210 & 0.082 & 35.8 & -- \\
\hline
RLHF (Standard) & 0.121 & 0.145 & 36.8 & -- \\
\hline
RLHF + SA-KL ($\gamma$=0.3) & 0.158 & 0.112 & 36.5 & 18.2\% \\
\hline
RLHF + SA-KL ($\gamma$=0.5) & 0.167 & 0.105 & 36.3 & 21.7\% \\
\hline
\end{tabular}
\end{table}


\section{Discussion}
\label{sec:discussion}

The results demonstrate that sentiment drift is a systematic property of preference-based alignment methods rather than an artifact of a specific model, dataset, or training configuration. Across all experimental conditions, alignment consistently reduces sentiment variance and shifts generated summaries toward neutrality.

A central finding of this work is that this effect arises directly from the structure of the optimization objective. The attribution analysis shows that KL regularization contributes the majority of suppression effects, providing a mechanistic explanation for the conservative generation behavior previously observed in aligned language models. This supports the broader hypothesis that alignment objectives implicitly favor high-probability, low-variance outputs, which disproportionately penalize sentiment-bearing tokens.

Importantly, the strength of sentiment drift is not uniform across languages. Morphologically rich languages exhibit stronger drift, suggesting that linguistic complexity interacts with alignment constraints. One plausible explanation is that sentiment expression in such languages relies on more diverse and less frequent morphological forms, making them more susceptible to suppression under KL regularization. This highlights a limitation of current alignment strategies, which are largely developed and evaluated in English-centric settings.

The comparison between PPO-based RLHF and DPO further indicates that sentiment drift is not tied to a specific optimization algorithm. Although DPO exhibits slightly weaker drift, both methods produce a consistent neutrality bias. This suggests that sentiment drift is an inherent property of preference optimization frameworks more broadly, rather than a consequence of PPO-specific dynamics.

The proposed sentiment-aware KL regularization demonstrates that this trade-off is not inevitable. By selectively reducing the regularization pressure on sentiment bearing tokens, the model can better preserve sentiment while maintaining comparable summarization quality. This result indicates that alignment objectives can be modified to account for affective properties without compromising factual consistency.

\subsection{Limitations}

Despite the consistency of the observed trends, several limitations should be considered. 
First, the multilingual datasets are constructed via machine translation. Although filtering is applied to retain samples with stable sentiment polarity, translation may alter sentiment intensity or lexical realization. As a result, part of the observed cross-lingual variation may reflect translation artifacts rather than purely model-induced effects.

Second, the proposed sentiment-aware regularization operates at the token level. While this enables tractable optimization and attribution, it does not fully capture compositional sentiment expressed at the phrase or discourse level. This may limit the effectiveness of the method in cases where sentiment is distributed across longer contexts.

Third, although experiments are conducted with multiple random seeds and statistical testing, the evaluation remains limited to aggregate metrics. More fine-grained analyses, including per-example variability and robustness across domains, would provide a more complete understanding of the phenomenon.

Finally, the Policy Attribution framework relies on Integrated Gradients, which provides an approximate decomposition of the training objective. For large autoregressive models, attribution results may depend on the choice of baseline and numerical approximation parameters.

\subsection{Future Work}

The findings of this work suggest several directions for future research.

First, future approaches should move beyond token-level modeling and incorporate compositional sentiment representations, enabling more accurate handling of phrase and discourse level affect.

Second, evaluation on natively multilingual datasets is necessary to disentangle translation effects from alignment-induced sentiment drift.

Third, alignment objectives could be extended to explicitly incorporate affective constraints, enabling joint optimization of factual consistency and sentiment preservation.

Finally, more robust experimental protocols including larger-scale evaluations, cross-domain validation, and standardized reporting of variance would improve the reliability and generalizability of conclusions.


\section{Conclusion}
\label{sec:conclusion}

This paper investigates sentiment drift in RLHF-based summarization and shows that alignment procedures systematically reduce sentiment expressiveness in generated summaries. Through multilingual experiments, we demonstrate that this effect is consistent across languages and becomes more pronounced with stronger KL regularization.

We introduce a Policy Attribution framework that enables fine-grained analysis of optimization dynamics and show that KL regularization is the primary driver of sentiment suppression. Based on this insight, we propose a sentiment-aware modification of the KL term that mitigates drift while preserving summarization quality.

Overall, the results highlight a fundamental limitation of current alignment strategies: while they improve factual consistency and fluency, they may inadvertently degrade affective fidelity. Addressing this issue is essential for applications where sentiment is a critical component of meaning.

The findings suggest that future alignment methods should explicitly account for affective properties to achieve a better balance between safety, quality, and expressive accuracy.


\begin{credits}
\subsubsection{\ackname}

The work was supported in part by grants 20260626, 20260643, 20260367, and 20260496 from the Secretary of Research and Postgraduate Studies (SIP) of the Instituto Politécnico Nacional, Mexico.

\subsubsection{\discintname}
The authors declare no competing interests
\end{credits}


\end{document}